\documentclass{article}

\PassOptionsToPackage{numbers,sort&compress}{natbib}
\usepackage[final]{nips_2018}
\usepackage[utf8]{inputenc} 
\usepackage[T1]{fontenc}    
\usepackage{hyperref}       
\usepackage{url}            
\usepackage{booktabs}       
\usepackage{amsfonts}       
\usepackage{nicefrac}       
\usepackage{microtype}      
\usepackage{graphicx}
\usepackage[scaled=0.85]{beramono}
\usepackage[font=it]{caption} 
\usepackage{multirow}

\usepackage[dvipsnames]{xcolor}

\title{Generating equilibrium molecules with deep neural networks}

\author{
Niklas W. A. Gebauer \\
Machine Learning Group\\
Technische Universit\"at Berlin\\
10587 Berlin, Germany\\
\texttt{n.wa.gebauer@gmail.com}\\
\And
Michael Gastegger\\
Machine Learning Group\\
Technische Universit\"at Berlin\\
10587 Berlin, Germany\\
\texttt{michael.gastegger@tu-berlin.de}\\
\And
Kristof T. Sch\"utt\\
Machine Learning Group\\
Technische Universit\"at Berlin\\
10587 Berlin, Germany\\
\texttt{kristof.schuett@tu-berlin.de}\\
}

\begin{document}

\maketitle

\begin{abstract}
  Discovery of atomistic systems with desirable properties is a major challenge in chemistry and material science. Here we introduce a novel, autoregressive, convolutional deep neural network architecture that generates molecular equilibrium structures by sequentially placing atoms in three-dimensional space. The model estimates the joint probability over molecular configurations with tractable conditional probabilities which only depend on distances between atoms and their nuclear charges. It combines concepts from state-of-the-art atomistic neural networks with auto-regressive generative models for images and speech. We demonstrate that the architecture is capable of generating molecules close to equilibrium for constitutional isomers of C$_7$O$_2$H$_{10}$.
\end{abstract}

\section{Introduction}

The discovery of molecules and materials with certain desirable properties is crucial for a large variety of applications, including the design of drugs, catalysts, batteries and solar cells.
However, accurate calculations of quantum-chemical properties are computationally expensive even for moderately sized compounds.
Moreover, the space of all possible chemical compounds is incredibly large such that an exhaustive search with quantum chemistry methods is infeasible.
An increasing amount of research proposes the application of machine learning techniques to the chemical domain in order to accelerate molecular discovery.
Several approaches use recurrent neural networks to sequentially generate SMILES~\cite{smiles} strings -- a common representation of compounds -- in an autoregressive fashion. The generators are biased towards certain chemical or biological properties by either restricting the training data set~\cite{drugRNN,drugRNNgupta,drugRNN2} or guidance by a separately trained properties predictor in a reinforcement learning setting~\cite{drugRNNRL}.
Another line of work trains VAEs on SMILES representations in order to allow straightforward interpolation, perturbation and optimization in the learned continuous latent space~\cite{molVAE1,molVAE2,molVAE3,GVAE}.
A disadvantage of string representations and graph-based approaches is their neglect of information encoded in the interatomic distances, which can aid the learning problem by e.g. establishing physical relationships between various building blocks of the molecule.
Therefore, the most accurate approaches for the prediction of chemical properties are interatomic machine learning potentials that make use of the atom positions of systems~\cite{ref1,ref2,ref16,ref3,ref17,ref4,ref10,ref5,ref7,ref11,ref6}.
This poses a problem since the search of chemical compound space requires a method to generate candidate equilibrium structures in three-dimensional space without having to calculate energies.

In this work, we propose a generative deep neural network capable of generating new molecular equilibrium configurations, i.e. atoms placed in three-dimensional space, given a training data set of reference molecules.
%
%
%
We combine two recent advancements from the fields of generative models and atomistic neural networks. 
On the one hand, autoregressive deep convolutional neural architectures have successfully been applied in several domains, e.g. images~(PixelCNN~\cite{pixelRNN2016,pixelCNN2016}) and speech~(WaveNet~\cite{van2016wavenet}). 
These models estimate joint probability distributions, e.g. over pixels of an image, with products of conditional probabilities. 
We adopt this idea and propose a factorization of the distribution over molecular configurations that enables sequential generation of molecules by placing atoms sequentially. 
On the other hand, energy prediction with chemical accuracy has been reported for SchNet~\cite{schutt2017schnet,schutt2018schnet}, a deep convolutional neural model specifically designed to learn representations of atomistic systems.
It acts directly on the continuous space of molecular configurations, i.e. atom positions and their nuclear charges, and by design adheres to the invariance of molecules to translation, rotation, and atom indexing.
We adapt the SchNet architecture to fit the sequential generation framework.

The key contributions of this work are as follows:
\begin{itemize}
	\item We propose a tractable factorization of the continuous distribution over molecular configurations into conditional probabilities that allows for \textit{sequential sampling} of atom positions. 
	We formulate the conditional probabilities in terms of \textit{distances} between atoms in order to obey the rotational and translational invariance of molecules. 
	\item We propose a novel autoregressive generative neural network architecture to model the conditional probabilities over distances between atoms.
	It is based on the SchNet architecture for prediction of chemical properties of atomistic systems.
	\item We demonstrate that the proposed architecture is capable of generating molecular configurations of constitutional isomers of C$_7$O$_2$H$_{10}$ close to equilibrium which have not been included in the training set.
	Moreover, we have discovered and validated plausible equilibrium molecular structures that have not been included in the test set either.
\end{itemize}

\section{Tractable factorization of distribution over molecular configurations}
\begin{figure}[tb]
	\centering
	\includegraphics[width=1\linewidth]{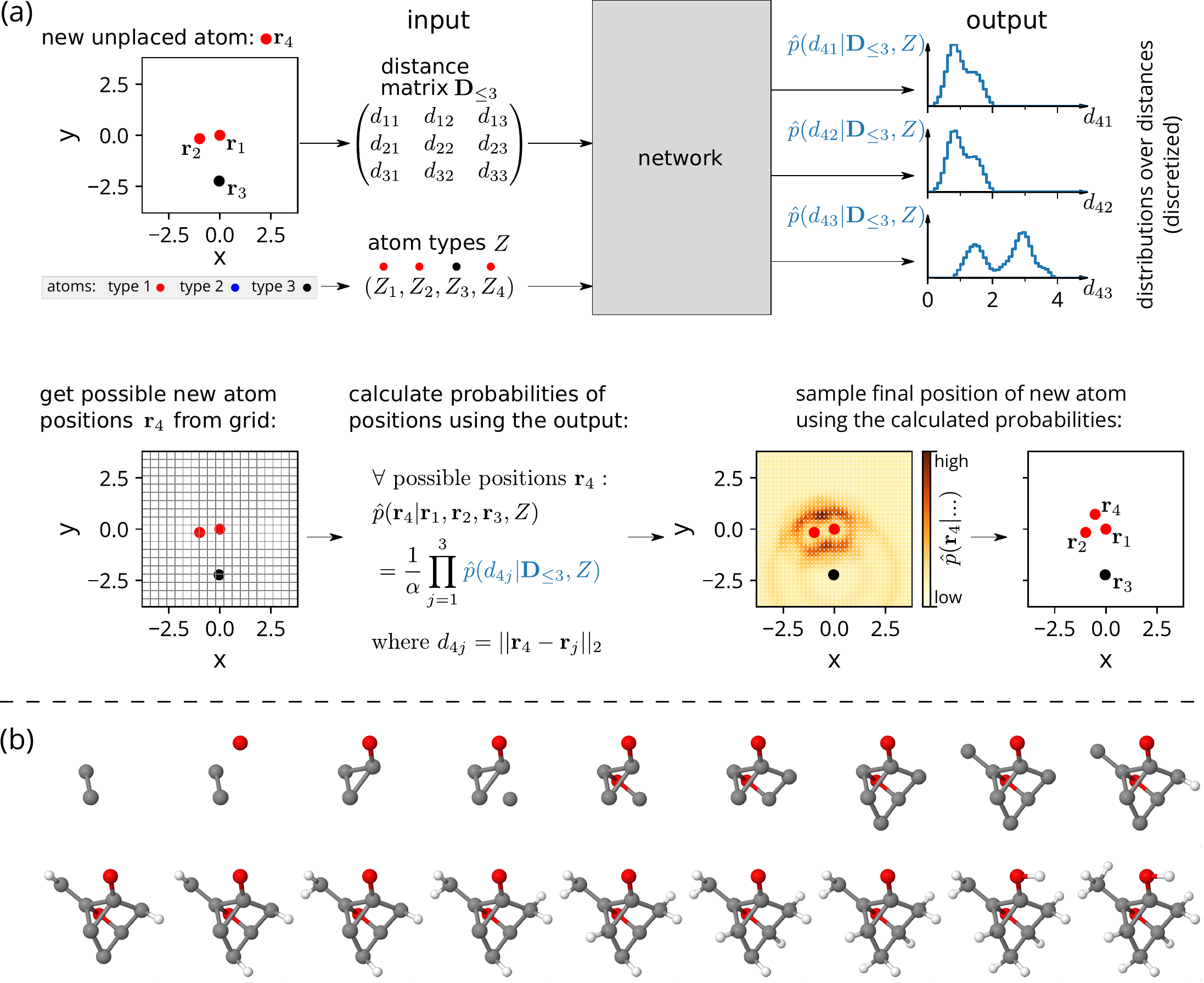}
	\caption[Scheme: Input/output of the network and how to generate a new absolute atom position]{Detailed depiction of a single step of the generation process using two-dimensional toy data (a). It shows the input and output of our architecture (top row) and steps taken when sampling the absolute position of a new unplaced fourth atom (bottom row). Starting from the point where two carbon atoms have been placed, the whole remaining placement process of a real-world C$_7$O$_2$H$_{10}$ isomer generated by our architecture is documented in (b).}
	\label{fig:generation scheme}
\end{figure}

Our aim is to generate new equilibrium molecules given a training data set of such molecules. 
To this end, we estimate the distribution over molecular configurations $p(m)$ with a tractable product of conditional probabilities. 
The generation process is illustrated in Figure~\ref{fig:generation scheme}.

Let $n$ be the number of atoms in a molecule. 
Then we define $p(m)$ as the joint distribution $p(R|Z)$ over all atom positions $R=(\mathbf{r}_1,...,\mathbf{r}_n)$ given nuclear charges $Z=(Z_1,...,Z_n)$ and propose the following factorization:
\[
p(m) = p(R|Z) = p(\mathbf{r}_1,...,\mathbf{r}_n|Z_1,...,Z_n) =
\prod_{i=1}^{n-1} p(\mathbf{r}_{i+1}|\mathbf{r}_1,...,\mathbf{r}_i,Z_1,...,Z_{i+1}).
\]
Note that this factorization allows to sample the positions of atoms in a molecule sequentially as the position of any atom $\mathbf{r}_{i+1}$ only depends on the atom's nuclear charge $Z_{i+1}$ and the positions and nuclear charges of all preceding atoms. 
Thus it can be used to train an autoregressive neural network to generate molecules atom by atom if nuclear charges $Z$ are provided. 

However, learning a distribution over atom positions poses multiple challenges.
Most importantly, molecular energies are invariant to rotation and translation.
Thus, the generative model should include these invariances for an efficient use of the available data.
Furthermore, working with absolute positions requires discretization with a 3-d grid that impairs scalability. 
To overcome these problems, we instead introduce a factorization that relies only on distances between atoms to describe the probability of a certain absolute position:
\[
p(\mathbf{r}_{i+1}|\mathbf{r}_1,...,\mathbf{r}_{i},Z_1,...,Z_{i+1}) = \frac{1}{\alpha} \prod_{j=1}^{i} p(d_{(i+1)j}| \mathbf{D}_{\leq i},Z_1,...,Z_{i+1}).
\label{eq:distr_pos}
\]
Here $\alpha$ is a normalization factor and $\mathbf{D}_{\leq i}$ is a matrix containing the distances between all positions $\mathbf{r}_1,...,\mathbf{r}_i$ of already placed atoms. 
The probability of a new atom position $\mathbf{r}_{i+1}$ is given as a product of probabilities of distances $d_{(i+1)j}$ between the new atom position and the positions of all preceding atoms.
Our architecture learns these distributions over distances instead of working with absolute positions directly.
It adheres to the invariance of molecules to rotation and translation by design as the modeled distributions only depend on nuclear charges $Z_1, ..., Z_{i+1}$ and distances $\mathbf{D}_{\leq i}$ of preceding atoms. 
This approach improves the scalability of our model as we are able to discretize distances in one dimension independent from the dimensionality of the underlying positions.
Using Eq.~\ref{eq:distr_pos}, we are able to calculate the probability of absolute atom positions.
While the generation process is sequential, the model can be trained efficiently in parallel, where the distances between atoms in the training data can be used directly as targets.

\section{Adapted SchNet architecture}
\begin{figure}[tb]
	\centering
	\includegraphics[width=1\linewidth]{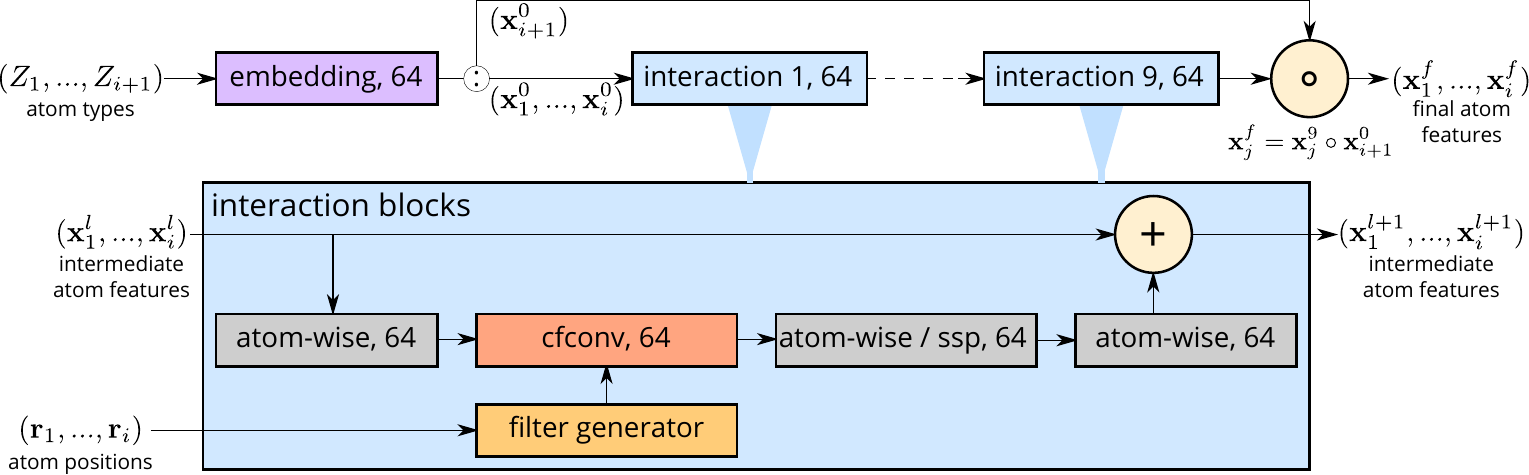}
	\caption[Scheme: Adapted SchNet architecture for feature extraction]{Feature extraction part of the adapted SchNet architecture made up of nine interaction blocks with dense atom-wise layers, shifted-softplus non-linearities (ssp), continuous-filter convolutions (cfconv) and filter generating layers as explained by Sch\"utt et al.~\cite{schutt2017schnet}. It differs in the split of features after the atom type embedding and concerning the entry-wise product after the last interaction block which provides feature vectors with information about the type of the new atom.}
	\label{fig:feature extraction scheme}
\end{figure}
The feature extraction of our autoregressive architecture is shown in Figure~\ref{fig:feature extraction scheme}. 
It is similar to SchNet~\citep{schutt2017schnet, schutt2018schnet} for the prediction of molecular properties. 
The embedding characterizing the atom types is split into feature vector $\mathbf{x}_{i+1}^0$ of the new atom $i+1$ and feature vectors $(\mathbf{x}_1^0,...,\mathbf{x}_i^0)$ of all preceding atoms. 
Here lays the main difference to the predictive SchNet architecture which always has access to the complete molecule. 
In contrast, our architecture works with partial molecular data, namely the positions $\mathbf{r}_1,...,\mathbf{r}_i$ of already placed atoms, their nuclear charges $Z_1, ..., Z_{1}$, and the nuclear charge $Z_{i+1}$ of an unplaced, new atom whose position $\mathbf{r}_{i+1}$ shall be sampled using the output of our network. 
The information about already placed atoms is processed just as in the predictive SchNet model, using interaction blocks to update feature vectors depending on the molecular geometry of the neighborhood. 
Additionally, the embedding of the nuclear charge of the atom to be placed is used to update the calculated feature vectors of preceding atoms with information about the new atom as a last step.
This is reflected by the element-wise product, which is written as "$\circ$" in Figure~\ref{fig:feature extraction scheme}.

\begin{figure}
	\centering
	\includegraphics[width=1\linewidth]{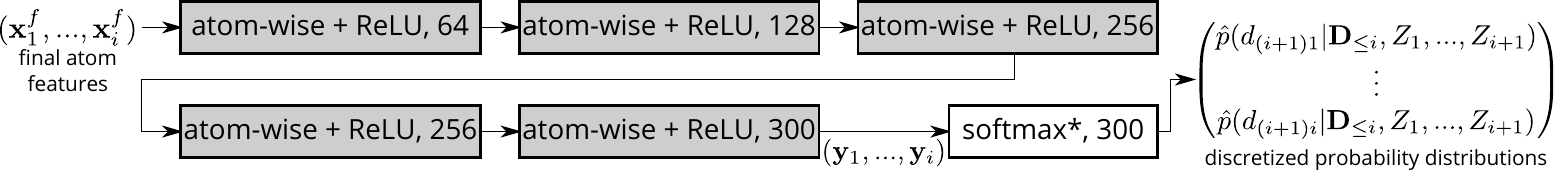}
	\caption[Scheme: Second part of architecture for calculation of probability distributions]{Second part of our autoregressive SchNet architecture which uses the extracted features of already placed atoms to calculate probability distributions over distances to the new atom.}
	\label{fig:pdf scheme}
\end{figure}
In this way, the architecture extracts one feature vector for every already placed atom. 
As shown in Figure~\ref{fig:pdf scheme}, these feature vectors are processed independently by several dense layers with increasing size to obtain the desired distributions over distances between every preceding atom and the new atom. 
We use vectors of length 300 as discrete representations of these distributions over distances in a range from zero to a maximum distance $d^{\max}$. Every entry then is a bin for the probability of a small distance range depending on the fixed maximum distance $d^{\max}$ which can be determined by evaluation of the distances in the data set. In our experiments the maximum distance was roughly 15{\AA} and accordingly the bins covered distances of $\sim$0.05{\AA}.
The discretized distributions are obtained by applying a softmax with temperature term to the network output vectors $\mathbf{y}_1,...,\mathbf{y}_i$:
\[
\hat{p}(d_{(i+1)j}|\mathbf{D}_{\leq i}, Z_1, ..., Z_{i+1}) = \frac{1}{\beta}\exp\left(\frac{\mathbf{y}_j}{T}\right).
\]
Here $\beta$ is a normalization constant obtained by summing over all entries of the vector after application of the exponential function. 
The temperature $T$ corresponds to the thermodynamic temperature in Boltzmann distributions.
High temperatures lead to more uniform, flat distributions whereas low values produce pronounced, peaked distributions. 
This is required to control the randomness during the molecule generation process but remains fixed at $T=1$ during training.

To obtain labels $\mathbf{q}_{(i+1)j}$ (where $j\leq i$) for predicted distributions $\hat{p}(d_{(i+1)j}|\mathbf{D}_{\leq i}, Z_1, ..., Z_{i+1})$, we calculate the true distances $d_{(i+1)j}$ between the supposedly new atom and the preceding atoms from a training molecule. 
The distances are expanded using Gaussian radial basis functions (RBF) with 300 linearly spaced centers $0\leq\mu_c\leq d^{\max}$ to obtain a discretized, one-dimensional, uni-modal distribution peaked at the label vector entry $c$ where the center is closest to the true distance:
\[
	\left[\mathbf{q}_{(i+1)j}\right]_c = \frac{\exp(-\gamma (d_{(i+1)j} - \mu_c)^2)}{\sum_{c=1}^{300} \exp(-\gamma (d_{(i+1)j} - \mu_c)^2)}.
\]
The width of the approximated distribution can be tuned with parameter $\gamma$. 
In our experiments we set it to the reciprocal of the distance between two neighboring centers.

To optimize our neural network, we make use of the cross-entropy between label vectors $\mathbf{q}_{(i+1)j}$ and network outputs $\hat{\mathbf{p}}_{(i+1)j} = \hat{p}(d_{(i+1)j}|\mathbf{D}_{\leq i}, Z_1, ..., Z_{i+1})$:
\[
	H\left(\mathbf{q}_{(i+1)j}, \hat{\mathbf{p}}_{(i+1)j}\right)= - \sum_{c=1}^{300} \left[\mathbf{q}_{(i+1)j}\right]_c \cdot \log\left[\hat{\mathbf{p}}_{(i+1)j}\right]_c.
\]
The loss is then given as the sum over the cross-entropy for all predictions made during the generation process of a complete molecule. 
If it has $n$ atoms, this means that $n-1$ atom positions need to be sampled, namely all but the position of the first atom. 
At each sampling step $i$ ($1\leq i < n$), the network provides $i$ distributions over distances, one for every already placed atom. 
Let $\hat{P}$ and $Q$ be the sets of all predicted distributions made during that process and the corresponding label vectors, respectively. 
Then the loss is defined as:
\[
	\ell(\hat{P}, Q) = \sum_{i = 1}^{n-1} \sum_{j=1}^{i} H\left(\mathbf{q}_{(i+1)j}, \hat{\mathbf{p}}_{(i+1)j}\right).
\]

\section{Experiments and results}

We train our generative architecture on the equilibrium structures of the 6,095 constitutional isomers of C$_7$O$_2$H$_{10}$ provided in the QM9~\cite{ramakrishnan2014quantum,dataset2/ci300415d} data set.
We use only the atom positions and nuclear charges of 2,000 of the C$_7$O$_2$H$_{10}$ isomers as training data and reserve the remaining 4,095 molecules as test data. 
The same maximum distance $d^{\max}$ is chosen for the RBF expansions in the filter-generating layers as well as for label generation. 
It is determined by the largest distance on the three-dimensional grid used for molecule generation, which extends from -$4.4${\AA} to $4.4${\AA} with 45 linearly spaced steps of length $0.2${\AA} in all three dimensions. 
Therefore, the maximum distance is given as the length of the diagonal connecting two corners of the cube: $d^{\max} = \sqrt{3\cdot8.8^2} \approx 15.24${\AA}. 
The total number of possible new atom positions on the grid is $45^3=91125$. 
The softmax temperature term that tweaks the form of distributions is $T=1$ during optimization but set to $T=0.01$ in order to increase precision of possible new atom positions and decrease randomness whenever molecules are generated.

We use randomly sampled mini-batches of size 20 where we permute the order of atoms randomly but make sure that the ten hydrogen atoms are always placed after the C$_7$O$_2$ substructure has been established.
We found that this simplifies learning as the placement of hydrogen atoms is principally deterministic given a proper C$_7$O$_2$ scaffold. 
The architecture is trained with an ADAM optimizer initialized with the standard hyper-parameter values suggested by \citet{kingma2014adam}. 
For validation, we generate 500 molecules every 1000 training iteration steps and predict their potential energy with a standard SchNet architecture as described by Sch\"utt et al.~\cite{schutt2017schnet} trained on energies and forces of perturbed configurations of 121 equilibrium C$_7$O$_2$H$_{10}$ structures from the ISO17~\cite{schutt2017schnet,schutt2017quantum,ramakrishnan2014quantum} data set. 
The sequential generation process uses randomized sequences of seven carbon and two oxygen atoms to build a C$_7$O$_2$ substructure and places ten hydrogen atoms afterwards. An example showing the sampling steps of one of the generated molecules can be found in Figure~\ref{fig:generation scheme} (b).

After training the architecture, we generate 10.000 C$_7$O$_2$H$_{10}$ structures. 
RDKit\footnote{RDKit: Open-source cheminformatics. \url{http://www.rdkit.org}} is used to compare fingerprints of generated molecules and data from the training and test sets. 
The fingerprints take into account all topological paths with lengths between one and seven bonds in the C$_7$O$_2$ substructures, while the placement of hydrogen atoms is not taken into account. 
According to this metric, the generator produced 4,392 unique molecular configurations including 266 structures that resemble training examples, but more importantly, 388 molecules that resemble unseen molecules from the test set.
%
\begin{figure}[tb]
	\centering
	\includegraphics[width=1\linewidth]{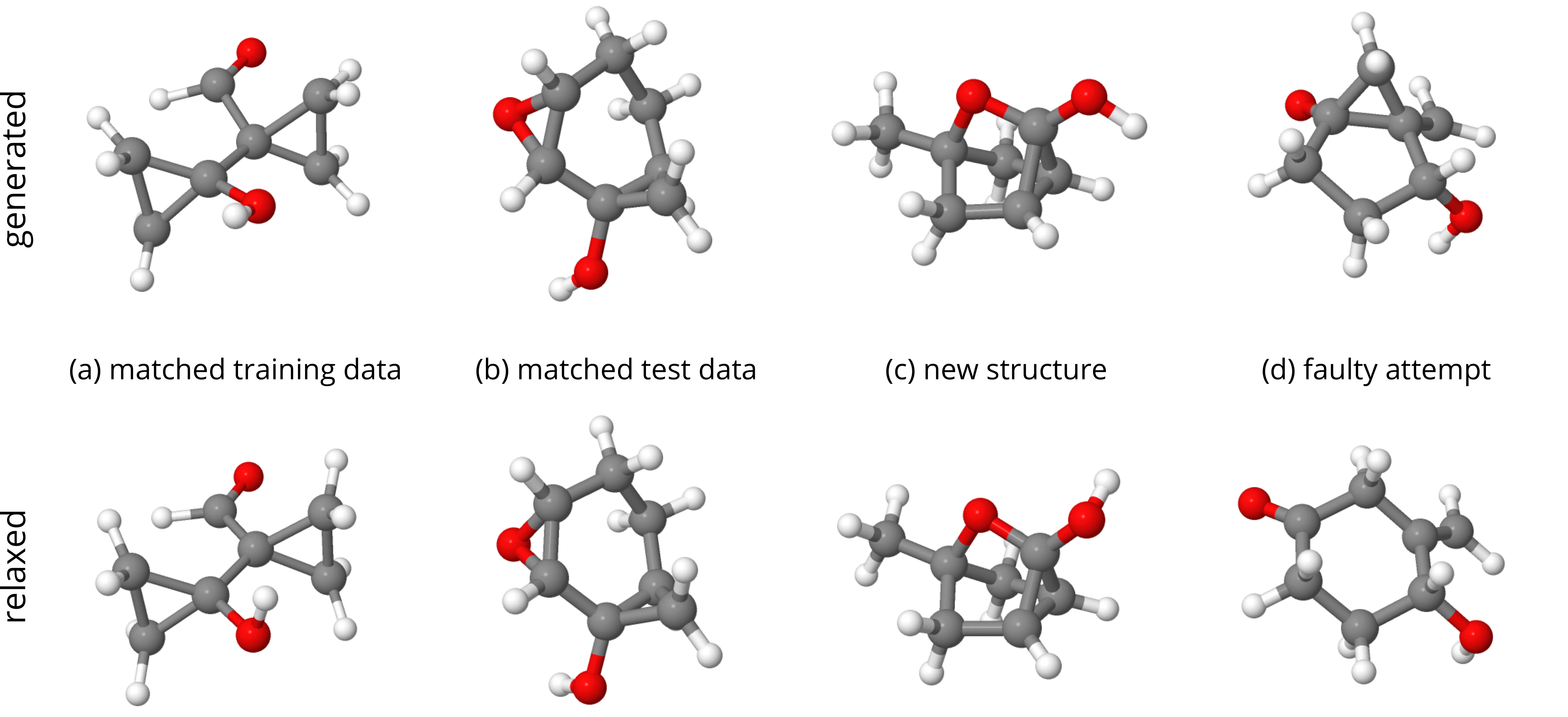}
	\caption[Examples of generated molecules]{Examples of generated molecules (top) and their configurations after relaxation (bottom) from the 20 best generation attempts according to predicted potential energy.}
	\label{fig:mols_gen}
\end{figure}

For further analysis, we select the 20 generated molecules with lowest predicted potential energy and optimize their structures at the PBE/def2-SVP level of theory using \textsc{Orca}~\cite{Perdew1996PRL,Weigend2005PCCP,Neese2012WCMS}. 
We plot four examples of the generated molecules with Jmol\footnote{Jmol: An open-source Java viewer for chemical structures in 3D. \url{http://www.jmol.org/}} before and after relaxation.
Figure~\ref{fig:mols_gen} shows one example that matched training data, one that matched test data, a valid new isomer that was neither in the test nor the training data as well as a faulty generation attempt where the relaxation significantly altered the molecular configuration. 
Overall, five of the tested 20 generated molecules matched training data, eight matched test data, two were completely new and five were failed attempts. 
Note that all failed attempts are structures where one or multiple atoms had a wrong number of covalent bonds. Therefore they could be identified prior to relaxation by checking the valence of the atoms and the number of covalent bonds in the generated molecules as a post-processing step.


\begin{table}
	\caption{Root-mean-square deviation (RMSD) of atom positions in {\AA} after relaxation for the 20 best generated molecules according to energy prediction, separately for 15 valid isomers and 5 failed attempts. For valid isomers, we separately report the contribution of generated molecules that match training data (5), test data (8) and which are new structures (2).}
	\label{tbl:rmsd}
	\centering
	\begin{tabular}{lrrrrr}
		\toprule
		&  \textbf{train} & \textbf{test} & \textbf{new}  & \textbf{all valid} & \textbf{failed} \\
		\midrule
		all atoms & 0.36 & 0.37 & 0.36 & 0.36 & 0.63\\
		heavy atoms & 0.20 & 0.20 & 0.17 & 0.20 & 0.53\\
		\bottomrule
	\end{tabular}
\end{table}
The relaxation only led to slight alterations of atom positions for the 15 successful attempts. 
The most notable changes usually were the orientations of O-H bonds as can be seen in examples (a) and (c). 
We used Open Babel~\cite{openbabel} to align the generated and relaxed molecules and determine the root-mean-square deviation (RMSD) of their absolute atom positions as a measure of change during relaxation. 
The results can be found in Table~\ref{tbl:rmsd}.
Compared to the five failed attempts, the RMSD for the 15 valid isomers was significantly lower considering all atoms (0.37{\AA} vs. 0.63{\AA}) as well as considering only heavy atoms (0.21{\AA} vs. 0.53{\AA}). 
Furthermore, it can be seen that the generated molecules corresponding to isomers from the training data did not change significantly less than examples which match test data or were completely new. 
Overall, the results show that the examined generated valid structures are in fact close to actual equilibrium configurations of C$_7$O$_2$H$_{10}$ isomers.

\section{Conclusions}

We have proposed a tractable factorization of the probability distribution over molecular configurations which only depends on distances between atoms. 
In this way, we were able to build an autoregressive, generative deep neural network architecture based on SchNet.
We have shown that it is capable of generating a variety of C$_7$O$_2$H$_{10}$ structures in three-dimensional space, which are geometrically close to actual equilibrium configurations. 
The generated molecules generalize to structures which were held back as test data as well as valid isomers which cannot be found among the 6095 molecules provided in QM9. 
This is a first step towards a generative model for molecules that directly provides chemical structures in three-dimensional space as opposed to SMILES strings or graph representations which require a subsequent search for corresponding stable molecular configurations.

Currently, the proposed approach depends on a model for energy prediction that was also trained for out-of-equilibrium configurations for the selection of candidate molecules.
Since the required training data will inevitably be produced during calculation of reference data and the validation of candidate molecules, we will aim to develop a unified architecture for generation and prediction that allows an integrated workflow of discovery and prediction.
In future work, we will furthermore extend the current architecture to molecules with varying composition as well as additional chemical properties for guided exploration of chemical compound space. 

\subsubsection*{Acknowledgments}
This work was supported by the Federal Ministry of Education and Research (BMBF) for the Berliner Zentrum für Maschinelles Lernen BZML (01IS18037A). MG acknowledges support provided by the European Union’s Horizon 2020 research and innovation program under the Marie Sk\l{}odowska-Curie grant agreement NO 792572.
Correspondence to NWAG and KTS.

\small
\bibliography{references}

\end{document}